# Assertion Enhanced Few-Shot Learning: Instructive Technique for Large Language Models to Generate Educational Explanations

TASMIA SHAHRIAR [0000-0003-0199-7757], KELLY RAMOS AND NOBORU MATSUDA [0000-0003-2344-1485]

NORTH CAROLINA STATE UNIVERSITY, RALEIGH NC 27695, USA

(TSHAHRI,KJFENNES, NOBORU.MATSUDA)@NCSU.EDU

Human educators possess an intrinsic ability to anticipate and seek educational explanations from students, which drives them to pose thought-provoking questions when students cannot articulate these explanations independently. We aim to imbue Intelligent Tutoring Systems with this ability using few-shot learning capability of Large Language Models. Our work proposes a novel prompting technique, *Assertion Enhanced Few-Shot Learning*, to facilitate the generation of accurate, detailed oriented educational explanations. Our central hypothesis is that, in educational domain, few-shot demonstrations are necessary but not a sufficient condition for quality explanation generation. We conducted a study involving 12 in-service teachers, comparing our approach to Traditional Few-Shot Learning. The results show that Assertion Enhanced Few-Shot Learning improves explanation accuracy by 15% and yields higher-quality explanations, as evaluated by teachers. We also conduct a qualitative ablation study to factor the impact of assertions to provide educator-friendly prompting guidelines for generating explanations in their domain of interest.

## 1 INTRODUCTION

Students benefit from providing correct and detailed explanations of their actions, deducing new knowledge, and recognizing and repairing their own error or misconceptions while working on a problem [1-3]. These explanations are defined as *knowledge-building* explanations. Researchers argue that students learn more when they provide knowledge-building explanations in response to questions [4, 5]. Most human teachers have an innate sense of a knowledge-building explanation that they anticipate from their students in a classroom [6]. This expectation drives them to ask follow-up questions when students' responses are vague, underspecified, ambiguous, fragmentary, and error-ridden [7]. Unlike human teachers, virtual agents embedded in Intelligent Tutoring Systems (ITS) still require huge domain-specific data, manual coding, and carefully curated curriculum scripts to comprehend the intricacies of knowledge-building explanations to generate thought-provoking follow-up questions [8, 9]. Automating the generation of knowledge-building explanations is thus a crucial step in enabling the virtual agents to ask thought-provoking follow-up questions.

The emergence of Large Language Models (LLMs) has garnered significant attention, particularly due to their capability to perform a new task by simply conditioning on a few input-output pairs (demonstrations) without any need of fine-tuning the model; this type of inference is called *in-context learning* or *few-shot learning* [10-13]. Previous research has showcased instances where LLMs conditioning on demonstrations have not only rivaled but even surpassed state-of-the-art performance benchmarks of fine-tuned LLMs in text classification, multi-choice task [14], open domain question answering task, translation task [10], arithmetic reasoning, commonsense reasoning and symbolic reasoning [15, 16].

However, the few-shot performance of LLM is susceptible to the textual content provided in the demonstrations, known as *prompt* [17-20]. This challenge becomes prominent when crafting the prompt for explanation generation tasks within educational domains like physics, algebra, chemistry, and others. The complexity arises from the necessity to incorporate text that goes beyond conventional language comprehension, which involves integrating essential concepts and relationships between these concepts to facilitate precise and accurate reasoning. The aforementioned complexity is also corroborated by the research of Rae et al. [21]. Their findings revealed that LLMs' few-shot performance often falls short in tasks demanding advanced reasoning abilities and does not improve substantially with increasing model scales.

These collective findings have spurred our interest in designing a prompt for generating knowledge-building explanations for educational domain that ensures the accurate incorporation of concepts while mitigating the risk of generating explanations that appear nonsensical or incoherent, a phenomenon referred to as *hallucination* [22]. We emphasize to streamline the prompt to a degree that allows educators from diverse domains to readily apply it, effectively reducing barriers for non-experts. Our central assumption is that demonstrations, while necessary, are not sufficient to fully harness the few-shot performance of LLMs in the explanation generation task. Therefore, we introduce a novel approach called *Assertion Enhanced Few-Shot Learning*. Our proposed prompt for this approach consists of two components: (1) the design of suitable task demonstrations, similar to *Traditional Few-Shot Learning*, and (2) the design of domain-specific concept integration as assertions into the prompt separated from demonstrations to enhance the few-shot performance of LLM.

We conducted a comprehensive survey involving 12 in-service teachers to rigorously evaluate the quality of explanations generated using *Assertion Enhanced Few-shot Learning* vs. *Traditional Few-Shot Learning*. Our specific objective in developing this approach is to empower virtual agents to ask thought-provoking questions while students engage in problem-solving activities. To align with this objective, we have tailored our knowledge-building explanation generation task to cater to educational problem-solving scenarios. We explore the following research questions: **RQ1**: Does *Assertion Enhanced Few-Shot Learning* reduce the hallucination in LLM, as perceived by teachers? **RQ2**: Does *Assertion Enhanced Few-Shot Learning* enhance the quality of knowledge-building explanations generated by LLM, as perceived by teachers?

In this paper, we contribute: (1) a systematic design of prompt for knowledge-building explanation generation task in educational problem-solving domains, (2) a comprehensive survey study with in-service teachers to assess the quality of LLM-generated explanation. (3) a quantitative analysis of the teacher ratings. In doing so we found that our proposed approach significantly reduced hallucination and enhanced the quality of explanations compared to *Traditional Few-Shot Learning*. (4) a qualitative ablation study to delve deeper into specific aspects of our approach, yielding comprehensive prompt design guidelines for knowledge-building explanation generation using LLMs.

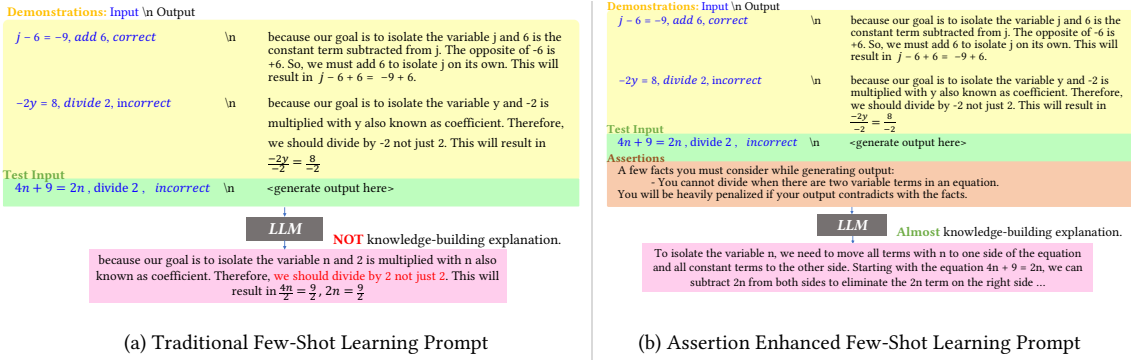

(a) Traditional Few-Shot Learning Prompt

(b) Assertion Enhanced Few-Shot Learning Prompt

Figure 1: The differences between Traditional vs Assertion Enhanced Few-Shot Learning Prompt

## 2 RELATED WORK

### 2.1 Zero-Shot Learning strategy for explanation generation task

Axelsson et al. [23] proposed a method of text generation from a knowledge-graph using Zero-Shot Learning. Their proposed prompt does not contain any demonstration of the task; it only contains the textual description or invocation of the task to be performed. A comparable approach to address our specific task involves formulating a prompt that defines a higher-level abstraction of the knowledge-building explanation within the problem-solving domain. The major drawbacks of such a prompt are: (1) we have almost no control over the pattern of the generated explanation (2) there is an elevated risk of producing scientifically inaccurate explanations, a phenomenon widely known as *hallucinations*. (3) one- and few-shot performance of LLM is often much higher than its zero-shot performance [10].

Our research distinguishes itself from [23] as we examine the few-shot ability of LLM, which allows for more precise control over the generated text. Furthermore, we explore modifications to the few-shot prompt aimed at mitigating or possibly reducing the occurrence of hallucinations within the educational problem-solving domain, all without the need for extensive fine-tuning efforts.

### 2.2 Traditional Few-Shot Learning with Chain-of-thought

Wei et al. [15] found that enhancing the accuracy of Traditional Few-Shot Learning is achievable by incorporating a human-like, step-by-step approach to task execution as natural language input within the demonstrations. To illustrate, instead of merely presenting math word problems and their corresponding final answers in the demonstrations, providing the accompanying thought process for solving the math problems in the prompt proves to be instrumental in unlocking the reasoning capabilities of LLMs. It is worth noting that the intermediate thought process expressed in natural language is essentially what we can offer as knowledge-building explanations. However, a recent study found that the existence of patterns in the chain-of-thought reasoning is crucial to its success. They claimed that chain-of-thought does not facilitate the reasoning capabilities of LLM; instead, the rationales merely act as a beacon for the model to realize "what" tokens to replicate in the output to form a factual answer [16]. Therefore, introducing different concepts and their relationships that drive the thought process across demonstrations in the educational domain can disrupt the established patterns in the chain-of-thought.

Informed by these two influential studies, our work aims to design the demonstrations with chain-of-thought, ensuring the preservation of the symbiosis between text and patterns. To do that, we have separated the illustrations of different concepts and their relationships as assertions in the prompt to "trick" the model into forming sentences that resemble correct answers in the educational domain.

## 3 TASK DESCRIPTION

In this section, we provide the formal definition of a knowledge-building explanation, drawing from established theoretical frameworks [24, 25]. During problem-solving, knowledge-building explanations are explanations that accurately explains why an action is correct or incorrect in a detailed oriented way. It must include essential conceptual terms that enrich the understanding. Furthermore, the explanation should reflect a genuine comprehension of the concept, demonstrating how it is effectively utilized within the explanation. In summary, a knowledge-building explanation comprises the following integral components:

1. The explanation is logically correct.
2. The explanation is relevant the action.
3. The explanation mentions the important concepts to make sense of the action.
4. The explanation reflects a genuine comprehension of the concept based on how it is utilized within the explanation.

Given a problem-solving scenario, we expect LLM to produce knowledge-building explanations related to any action a student may perform in an attempt to solve the problem. If the performed action is correct (or incorrect), LLM is expected to provide an accurate, detailed and concept-oriented explanation discussing why the action is correct (or incorrect).

## 4 ASSERTION ENHANCED FEW-SHOT LEARNING

This section outlines our approach to design the *Assertion Enhanced Few-Shot Learning* prompt. Our prompt design involves two stages: (1) First, we create suitable demonstrations of the task akin to a *Traditional Few-Shot Learning* prompt and, (2) Second, the incremental process of integrating the domain-specific assertions into the prompt. While we primarily discuss the linear algebraic equation-solving domain for simplicity, our approach is adaptable to any problem-solving domains.

### 4.1 Design of Demonstrations

Demonstrations are the list of input-output pairs conditioned on which LLMs learn about the task to be performed. For instance, if an LLM is expected to translate a test English sentence to Bengali, the demonstrations include a few examples of English sentences mapped to their corresponding Bengali translation. The test English sentence is added at the end of the demonstrations, and the LLM is expected to generate the Bengali translation. Here, each English sentence is defined as *input* and the corresponding Bengali translation as *output* for each demonstration. In this section, we discuss the structure of the input and output pair for knowledge-building explanation generation task.

*4.1.1 Structure of input in a demonstration*

In classification tasks, the input distribution across all demonstrations significantly impacts the *few-shot* performance of LLMs [14]. Therefore, inputs in our task must capture all problem-solving scenarios that could influence the generation of knowledge-building explanations. Therefore, we defined input consisting of three components: <*problem state, solution step, correctness of the solution step*>. We hypothesized that these three components are sufficient to capture the input distribution of the problem-solving domain, and the simplicity of the components makes it easier for educators to effortlessly prompt the LLM to generate knowledge-building explanations within their domain of interest, reducing entry barriers.

*Problem state* can be described as the snapshot of a scenario that a student would be asked to solve. It consists of all the pertinent details and conditions that define the situation at hand. This can range from straightforward instances like $125\ g\ =\ ?\ oz$ in the unit-conversion domain to more intricate challenges such as a physics problem, "You have a ball, and you drop it from a height of 10 meters above the ground. Ignoring air resistance, your task is to find out how long it takes for the ball to hit the ground and how fast it's moving just before it hits." In our chosen domain of algebraic equation solving, problem states encompass various types of equations, for instance: $j - 6 = 9$.

*Solution step* can be defined either as the final answer of a problem state or any intermediate step that would simplify the problem state depending on which level educators want their virtual agent to engage with the students during problem-solving. For instance, in unit-conversion domain, we can simply include the final answers like 4.41 oz as solution, whereas, in algebraic equation solving, we can include transformations like performing *add, subtract, divide, or multiply* by *a term* on both sides of the problem state as solution step.

*Correctness of the solution step* is defined as the Boolean indicator of whether the mentioned solution step is correct or incorrect. This component allows the inclusion of both accurate and erroneous steps that students might present when tackling a problem state.

Figure 1 shows the construction of *input* highlighted in blue text for algebraic equation solving domain. In our design of *input*, we chose to retain the numerical quantities as placeholders for both the problem state and solution steps, instead of substituting them with generalized symbols. For instance, we defined input as <problem State: $j - 6 = -9$, solution Step: $add\ 6$, correctness of the solution step: correct> instead of <problem state: $v - N_1 = N_2$, solution step: $add\ N_1$, correctness of the solution step: correct>. This decision was informed by a study that concluded that the specific values and symbols are mostly immaterial to the model performance [16]. Consequently, our hypothesis was that maintaining the problem state and solution step in their original numerical form would result in a more intuitive design.

*4.1.2 Structure of output in a demonstration*

Output is the part of a demonstration which is representative of the task itself. For the knowledge-building explanation generation task, the output is an elaborated, in-depth reasoning in favor or disfavor of a solution step using important concepts of that domain. However, these explanation should follow a consistent pattern across all demonstrations for the effective functioning of LLM reasoning capability [16]. If we introduce and explain varying concepts for unique problem and solution step within a single prompt, the induced "noise" from demonstrations creates a virtually equivalent variant of having no pattern at all, hurting the reasoning capability of LLM [16].

This constraint limits what we can put into the output. We hypothesize that when explaining the steps, the procedural aspects mentioning the relevant concept terms tend to follow a pattern, and we can include this pattern as part of the output .However, we argue that the conceptual definition of terms, for instance, the generalized rule when a concept applies or not require a different place in a prompt, which leads us to the formulation of domain-specific assertions (discussed in the subsequent section).

Figure 1 shows the sample output in the prompt. We only mentioned essential concepts in the output and avoided the definition of concepts in the output to ensure a consistent pattern across all demonstrations. It is important to note that the definition of these concept terms is not a property of knowledge-building explanation rather the appropriate usage of these concept terms is what constitutes as knowledge-building explanation as described in section 3.

**Number of demonstrations**: Finally, we selected eight input-output pairs as shown in the appendix, **Table 4** that are the most representative of our domain, as previous research [14, 20] has indicated that the model performance does not increase much after eight demonstrations. The resultant prompt is referred as Traditional Few-Shot Learning prompt throughout the paper and serves as our fundamental baseline prompt.

## 4.2 Design and Integration of Assertions

*Assertions* are defined as the conceptual knowledge of the domain that drives the procedural explanation of the inputs. For instance, "we divide by -2 when the equation is $-2y = 8$ because our goal is to isolate the variable y and -2 is multiplied with y also known as coefficient. Therefore, we should divide by -2 not just 2" is a procedural explanation and "the number multiplied with variable term is called the coefficient" is the conceptual knowledge aka an assertion. We hypothesized that including this conceptual knowledge of a domain in the demonstration itself may break the pattern across all demonstrations and therefore, must be placed at the end of few-demonstrations as an additional instruction. Figure 1 shows the placement of assertions in the prompt.

The inclusion of assertions followed an iterative process where we identified the conceptual error in the LLM generation. For instance, when initially using the traditional few-shot prompt, discussed in the previous section, one generation resulted in the following:

Problem state: $\frac{7}{3} = v + \frac{2}{3}$ Solution step: add $\frac{5}{3}$, correctness: incorrect, Knowledge-building explanation: because we cannot add a fraction directly with a variable v. We need to first multiply 3 to get rid of the fraction and isolate v on its own.

This generation prompted us to include the assertion, "we can add or subtract a fraction directly if the fraction is related to the variable term". With this assertion, the explanation evolved as follows:

Problem state: $\frac{7}{3} = v + \frac{2}{3}$ Solution step: add $\frac{5}{3}$, correctness: incorrect, Knowledge-building explanation: because our goal is to isolate the variable v and $\frac{2}{3}$ added with v, we want to do the opposite of $+\frac{2}{3}$ which is $-\frac{2}{3}$. So, we must subtract $\frac{2}{3}$ to isolate v on its own.

Similarly, we added other assertions through iterative identification of major conceptual gaps in the generations. This incremental approach can be viewed as a systematic method for instructing an LLM, focusing on rectifying generation errors progressively. **Table 5** in the appendix provides a summary of the assertions incorporated into our traditional few-shot prompt.

## 5 EXPERIMENT

To evaluate if the LLM generated explanations meet the criteria of knowledge-building on human discernment, we conducted a survey study with in-service teachers. The study aims to identify which prompt yields more knowledge-building explanations.

## 5.1 Stimuli

We selected the test input for explanation generation from the *cti_study_2022* dataset, accessed via DataShop [26]. We chose this dataset because it records middle school students' interaction with an Intelligent Tutoring System [27], where students teach the embedded virtual agent how to solve linear algebraic equation step-by-step by applying transformations on both sides. The dataset also captures the expert evaluation of the correctness of the solution step with respect to a given problem state. To ensure data quality and cost effectiveness, we removed all the duplicate input tuples from the dataset that share same values for all three components. The duplicate scenario in the dataset arises from two cases (1) multiple students working on the same problem state and suggesting the same solution step to the teachable agent and (2) same student teaching the same problem state and solution step to the agent multiple times. After this cleaning process, we randomly sampled 62 unique test input from the dataset. Each input is unique, as no two share the exact combination of components. For instance: $<3v + 2 = 7, add - 2, \text{correct}>$ and $<3v + 2 = 7, subtract\ 2, \text{correct}>$ are two unique inputs despite sharing the same problem state and correctness as they differ on the solution step.

We used the GPT-3 model, a state-of-the-art LLM to generate explanations for the test inputs employing both traditional few-shot prompt and assertion enhanced few-shot prompt. Consequently, each test input is associated with explanations generated by both prompts and these pre-generated explanations from LLM served as stimuli in this experiment.

## 5.2 Participants

We used purposive and snowball sampling to recruit participants. Recruitment inclusion criteria specified participants be currently teaching middle school math and have at least 3 years of experience teaching linear algebraic equation solving. A total of 47 in-service teachers were sent the link to our survey via email, among which only 12 in-service teachers could complete the study within the 10-day completion window. The survey itself had an approximate duration of two hours. Teachers received compensation upon completion at a rate of $30/h in the form of Amazon E-gift cards.

## 5.3 Procedure

First, teachers provided their consent to participate in the survey and for sharing the anonymous research data.

Second, to ensure that teachers understood the concept of knowledge-building explanation prior to performing the explanation evaluation task, they were given a one-page description of knowledge-building explanations in layman's terms with examples. They were also quizzed on two curated explanations outside our stimuli to validate their proper understanding. All 12 teachers passed the basic validation test.

Third, teachers entered the main task where they evaluated 20 stimuli in total, 10 randomly chosen from traditional few-shot prompt generation set and 10 randomly chosen from assertion enhanced few-shot prompt generation set. No teacher evaluated stimuli generated from both the prompts on the same inputs to mask out the input comprehensibility bias on the ratings. The teachers were presented with the test input, in random order, in a situated scenario, "A student and a virtual agent are arguing on performing the _solution step_ to simplify the _problem state_. The solution step is _correctness of the solution step_. The student provided the following explanation: _LLM-generated explanation_. Assess the following statements about the student's explanation." To maintain neutrality in the evaluation process, information regarding which prompt had generated the explanation was deliberately concealed using the situated scenario.

The metrics for evaluating the explanation was guided by theoretical definitions of knowledge-building discussed in section 3 and, we further refined the statements corresponding to those metrics through insights gained from the pilot study conducted before launching the main survey. The pilot study involved 6 in-service teachers whose results were not included during our analysis. We employed the think-aloud protocol to assess their perspectives on the following statements that use a five-point Likert scale (1 = strongly disagree, 5 = strongly agree):

1. The explanation is logically correct.
2. The explanation is relevant to the input.
3. The explanation mentions the important concepts to make sense of the solution step.
4. The explanation reflects a genuine comprehension of the concept based on how it is utilized within the explanation.

Our observations revealed that teachers exhibit a broad range of tolerance for logical correctness. For example, they tended to deem an explanation logically correct even if it supported a sub-optimal solution step, particularly in situations where student and virtual agent debated sub-optimal steps without an outright incorrect rationale. However, during few-shot demonstrations, *correctness of the solution step* was tagged as incorrect for sub-optimal solution steps, mimicking the expert evaluation of the solution step in the *cti_dataset_2022*. Therefore, we consider explanations pertaining to sub-optimal steps as logically inaccurate. To capture this nuance, we introduced an additional five-point Likert statement. Below are the statements included in the final survey for evaluating the explanation:

1. The explanation reflects that the explainer can simplify the equation state using optimal step.
2. The explanation is logically correct.
3. The explanation is relevant to the input.
4. The explanation mentions the important concepts to make sense of the solution step.
5. The explanation reflects a genuine comprehension of the concept based on how it is utilized within the explanation.

Finally, upon the completion of the survey, an automated ID was generated that the teachers emailed to the authors for compensation.

### 5.4 Measurements

Following the completion of the entire survey, we discarded two test inputs and their corresponding stimuli because the problem states for those stimuli were not parsed correctly into the survey due to some technical glitch. Two teachers evaluating those stimuli mentioned about the glitch in their response:

*"This survey question is messed up - we see "#NAME?" instead of an equation."*

*"Cannot answer the equation. Error in the given question."*

As a result, we excluded these two problematic inputs, resulting in a dataset of 60 inputs, each of which received at least one rating for explanations generated by both prompts from different teachers. For the stimuli that had more than one teacher rating were aggregated for each statement.

*5.4.1 Accuracy of the explanation.*

We calculated the accuracy of the explanations using teachers' rating on two statements: (1) explanation reflects that the explainer can simplify the equation state using optimal step and (2) explanation is logically correct. We created a new binary accuracy metric where correct corresponds to teacher saying *strongly agree* or *agree* to both the metrics, incorrect otherwise. This measure is purposefully made strict to screen out the slightest inaccuracy in the explanation that can be extremely crucial to learning in educational domains.

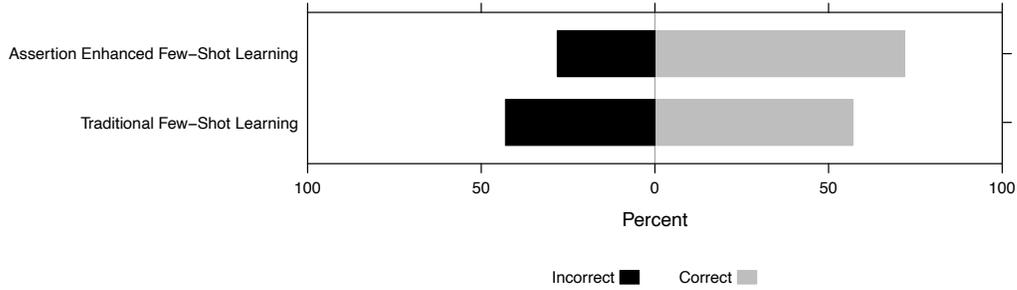

Figure 2: Teachers' rating of explanation accuracy. Traditional Few-Shot Learning generated accurate explanations 52% of the times. In-contrast, Assertion Enhanced Few-Shot Learning generated accurate explanations 67% of the times.

$$Accuracy = \begin{cases} 1, & \text{explains optimal step} = agree \mid\mid strongly\ agree \\ & \text{and} \\ & \text{explanation is logically correct} = agree \mid\mid strongly\ agree \\ 0, & \text{otherwise} \end{cases}$$

*5.4.2 Quality score of explanation.*

The overall quality score of an explanation was quantified as an average of accuracy and ratings on the rest three Likert-scale statements: (3) explanation is relevant to the input (4) explanation mentions the important concepts to make sense of the solution step (5) explanation reflects a genuine comprehension of the concept based on how it is utilized within the explanation. The quality score serves as a direct metric for assessing the knowledge-building potential of an explanation. Specifically, a higher quality score, derived from these comprehensive metrics, indicates a greater likelihood that the explanation qualifies as a knowledge-building explanation [24, 25].

## 6 RESULT

### 6.1 Traditional Few-Shot Learning generated accurate explanations 52% of the times

We investigated how often LLMs suffer from hallucinations while generating explanations using Traditional Few-Shot Learning. *Hallucination* refers to generating nonsensical or unfaithful text [28-30]. It is worth noting that human accuracy assessment remains the preferred method for evaluating hallucinations in Natural Language Generation compared to automatic evaluation approaches [31, 32]. To measure hallucination, we analyzed the explanation accuracy rated by the teachers. Figure 2 shows the distribution of the explanation accuracy. The data indicates that the probability of LLM generating accurate explanations using Traditional Few-Shot Learning is .52.

### 6.2 Assertion Enhanced Few-Shot Learning generated accurate explanations 67% of the times

We proceeded to investigate whether including assertions in the prompt contributed to a reduction in hallucination. The distributions of teacher rating between two prompts (shown in Figure 2) indicate that the probability of LLM generating accurate explanations using Assertion Enhanced Few-Shot Learning is .67. To confirm the statistical significance of this difference, we conducted a one-way ANOVA on the explanation accuracy with prompt as our independent variable. The result revealed a weak main effect for the prompt; $M_{\text{AssertionEnhanced-FS}}$ = .67±.47, $M_{\text{Traditional-FS}}$ = .52±.50, $F_{\text{Prompt}}(1, 118)$ = 2.81, $p$=0.09.

### 6.3 Assertion Enhanced Few-Shot learning has a higher teacher rating for focusing the explanation on the optimal step, relevant to the input, and mentioning the essential concepts to make sense of the solution step

We conducted a Multivariate Analysis of Variance (MANOVA) to assess how the two prompts impact teacher ratings across five metrics. The MANOVA result indicates a significant main effect of the prompt on three metrics: focusing the explanation on the optimal step ($M_{\text{AssertionEnhanced-FS}}$ = 4.03±1.31, $M_{\text{Traditional-FS}}$ = 3.48±1.64, $F_{\text{Prompt}}(1, 118)$ = 4.14, $p < .05$), relevancy to the input ($M_{\text{AssertionEnhanced-FS}}$ = 3.99±1.29, $M_{\text{Traditional-FS}}$ = 3.46±1.57, $F_{\text{Prompt}}(1, 118)$ = 4.23, $p < .05$) and mentioning the essential concepts to make sense of the solution step ($M_{\text{AssertionEnhanced-FS}}$ = 4.08±1.20, $M_{\text{Traditional-FS}}$ = 3.57±1.48, $F_{\text{Prompt}}(1, 118)$ = 4.41, $p < .05$). There is a weak main effect of the prompt on the metric reflecting a genuine comprehension of the concept based on how it is utilized within the explanation ($M_{\text{AssertionEnhanced-FS}}$ = 3.93±1.26, $M_{\text{Traditional-FS}}$ = 3.46±1.53, $F_{\text{Prompt}}(1, 118)$ = 3.44, $p = .07$). There is no main effect on the logical correctness of the explanation ($M_{\text{AssertionEnhanced-FS}}$ = 4.07±1.19, $M_{\text{Traditional-FS}}$ = 3.67±1.58, $F_{\text{Prompt}}(1, 118)$ = 2.45, $p = .12$). The top distribution in Figure 3 visually represents how teachers ratings vary between prompts for each metric.

These findings suggest that Assertion Enhanced Few-Shot Learning significantly outperformed Traditional Few-Shot Learning on generating explanations centered around an optimal step, being relevant to the input and mentioning essential concept terms while generating explanations. While this observation might appear straightforward in cases where correct solution steps were explicitly provided in the test input, as exemplified by this tuple <problem state: $-4n = 8 + n$, solution step: $subtract\ n$, correctness of the solution step: correct>. However, it becomes particularly noteworthy when considering scenarios where the test input includes an incorrect solution step, for instance, a tuple like <problem state: $5x = 3x - 2$, solution step: $divide\ 5$, correctness of the solution step: incorrect>.

We also conducted a one-way ANOVA on the quality score of explanation with prompt as our independent variable. The result revealed a weak main effect of prompt: $M_{\text{AssertionEnhanced-FS}}$ = 3.83±1.37, $M_{\text{Traditional-FS}}$ = 3.32±1.68, $F_{\text{Prompt}}(1, 118)$ = 3.22, $p$=0.07.

### 6.4 Assertion Enhanced Few-Shot Learning significantly improved the quality of explanations for the test inputs in which Traditional Few-Shot performed poorly

To analyze the variation in quality scores between prompts across varying test inputs, we divided the inputs into two groups based on the quality score of explanations generated by the traditional few-shot prompt. Subsequently, we categorized them as low complexity (when explanations by Traditional Few-Shot prompt scored above the median) and high complexity (when explanations by Traditional Few-Shot prompt scored below or equal to the median). Figure 4a shows the interaction plot with quality score as our dependent variable and prompt and Input complexity as our independent variable. The plot indicates that assertions notably improved the quality of explanations for input where the Traditional Few-Shot prompt generated below-average quality scores. Conversely, a slightly less favorable performance of Assertion Enhanced Few-Shot Learning was observed for inputs where the Traditional Few-Shot Learning generated above-average quality scores. To confirm if the interaction is significant, we performed a two-way ANOVA test with quality score as dependent variable and Prompt and Input Complexity as our independent variables. The result revealed a significant interaction between Prompt and Input Complexity: $F_{\text{Prompt:InputComplexity}}(1, 116)$ = 26.07, $p < .001$.

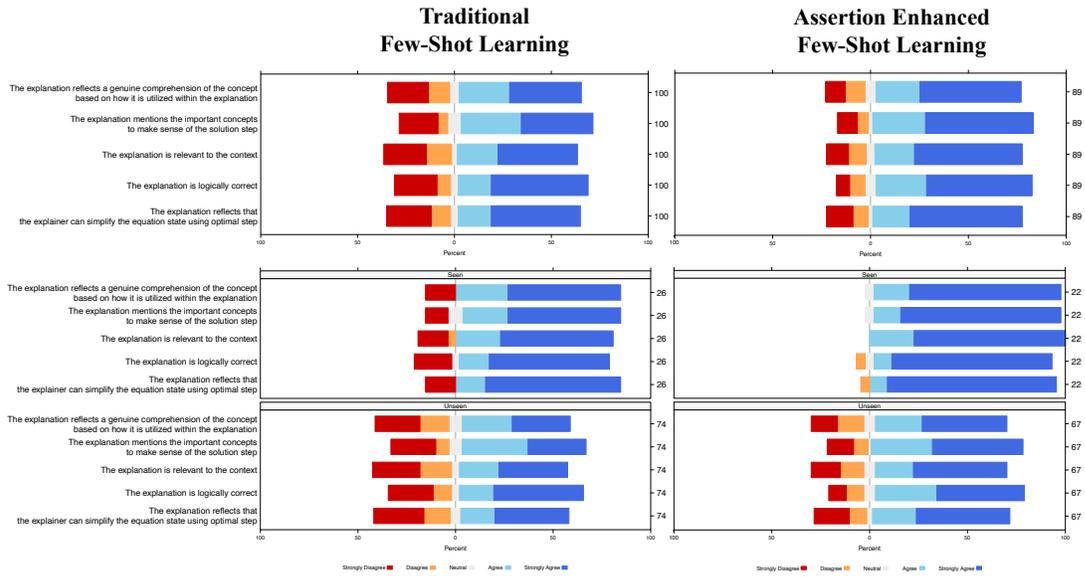

Figure 3: Distribution of teachers' ratings of five metrics between Traditional Few-Shot vs Assertion Enhanced Few-Shot prompt (top), further split between seen vs unseen inputs (bottom)

## 6.5 Traditional Few-Shot Learning mostly generated below average score for the unseen inputs and only a few assertions helped generating better quality response on those inputs without any need for new demonstrations in the prompt

We sought to understand which inputs were particularly complex for the prompts to generate quality explanations. We posed two hypotheses (1) the test inputs that closely resemble those *seen* in the demonstrations are the ones that are less complex to generate high quality explanation for both the prompts. (2) *unseen* test inputs are more complex, for which the Traditional Few-Shot Learning struggled to produce high-quality explanations.

To categorize inputs into seen and unseen, we defined our input space as follows:

Problem state: one step, two step with one variable term, two step with two variable terms, three step
Solution step: add +ve, add -ve, subtract +ve, subtract -ve, divide +ve, divide -ve, multiply +ve, multiply -ve
Correctness of solution step: correct, incorrect

We then automatically tagged both test and demonstration inputs using the space values. A test input is labeled as *seen* if it matches the values of problem state, solution step and correctness as the input value of the demonstrations, *unseen* otherwise. The tagging activity resulted in 13 seen and 47 unseen test inputs.

Figure 4b shows the scatterplot of quality score generated by Assertion Enhanced Few-Shot Learning vs quality score of Traditional Few-Shot Learning on *seen* and *unseen* inputs. The graph suggests that (1) most of the test inputs for which Traditional Few-Shot achieved below average score are unseen ones; for seen ones, the quality was mostly above average (2) for the unseen test inputs on which it performed poorly, Assertion Enhanced Few-Shot had better quality score as most of the scores lies above the 45-degree line. (3) Assertion Enhanced Few-Shot improved the quality score of *seen* test

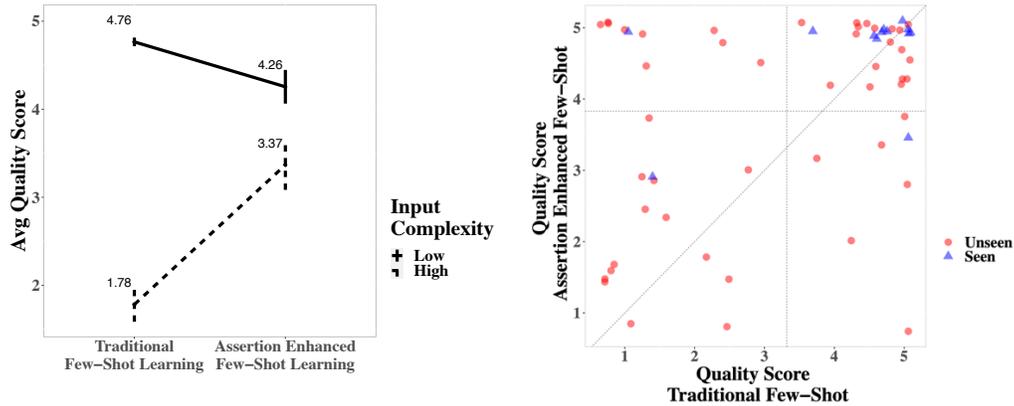

(a) Interaction plot: Quality score vs Prompts (Traditional vs Assertion Enhanced) and Input Complexity (High vs Low)

(b) Scatter plot: Quality score generated by Assertion Enhanced vs Traditional Few-Shot, color-coded for Seen and Unseen Inputs

Figure 4: Quality score comparison between prompts based on Input complexity

inputs even more, a trend also apparent from the teacher rating distribution in the bottom part of Figure 3 and (4) there are a few seen and unseen inputs located at the bottom right part of the scatterplot for which Assertion Enhanced Few-Shot performed poorly.

## 7 QUALITATIVE ABLATION STUDY

For additional ablation, we first randomly selected 5 test inputs for which Assertion Enhanced Few-Shot Learning generated high-quality explanations but Traditional Few-Shot performed poorly (shown in Table 1).

### 7.1 Is the value of adding assertions comparable to increasing the number of demonstrations?

1 out of 5 test inputs that we selected for the ablation study was previously marked as *seen* and 4 inputs were *unseen* label. To simulate the benefit of increased number of demonstrations, we added four new similar demonstrations to our Traditional Few-Shot prompt that match the 4 *unseen* inputs. We appended the new input-output pairs based on their problem type. We did not assume any specific order-related impact while appending the demonstrations. The newly added demonstrations are shown in appendix, **Table 6**.

To distinguish between the prompts, we named the Traditional Few-Shot prompt that we have discussed so far in our work as Traditional Few-Shot ($k = 8$) and the resultant prompt after adding four new demonstrations as Traditional Few-Shot with more demonstrations ($k = 12$), where $k$ represents the number of demonstrations. Table 1 shows the explanations generated by these two prompts. Our qualitative observation on how the explanations differ among Assertion Enhanced Few-Shot, Traditional Few-Shot ($k = 8$) and Traditional Few-Shot with more demonstrations ($k = 12$) are as follows:

5. Traditional Few-Shot ($k = 8$) could provide the optimal solution step for one out of the five inputs (test input 2), albeit with an incorrect line of reasoning. In contrast, Traditional Few-Shot with more demonstrations ($k = 12$) could not provide the optimal solution step for any of the five test inputs. Assertion Enhanced Few-Shot, on the other hand, could suggest the optimal solution step for three out of the five inputs with correct line of reasoning (test input 3-5).

6. The reasoning presented in the explanations differed between Traditional Few-Shot ($k$ = 8) and Traditional Few-Shot with more demonstrations ($k$ = 12), except for test input 1. They focused on different aspects of the problem state, although both provided incorrect explanations even for the sub-optimal solution steps.

The observations mentioned above did not demonstrate any discernible advantages of incorporating additional demonstrations. This observation aligns with prior research findings [14] that LLM performance does not increase much as $k$ increases.

### 7.2 Do assertions provide benefits similar to adding extra contextual information in the output?

We aimed to investigate whether the benefits of assertions could be attributed to the extra textual information in the prompt. Therefore, for each demonstration in the Traditional Few-Shot prompt ($k$ = 8), we appended relevant assertion statements to the output of each demonstration. Our hypothesis was that if assertions' benefits were solely due to this additional textual information, then adding all this extra text in the demonstration should result in explanations similar to those generated by Assertions Enhanced Few-Shot. As illustrated in Table 1, Traditional Few-Shot prompt with assertions embedded into demonstrations consistently failed to reach the optimal solution step and instead produced explanations containing irrelevant assertion-like statements, appearing more like an attempt to sound sophisticated than to provide meaningful explanations. These findings corroborate the essential role of patterns in facilitating chain-of-thought reasoning [16]. Moreover, they support our hypothesis that the inclusion of assertions within the demonstrations can be detrimental to the explanation generation process.

## 8   DISCUSSION AND PROMPT DESIGN GUIDELINES

### 8.1  RQ1: Does Assertion Enhanced Few-Shot Learning reduces the hallucination in LLM, as perceived by teachers?

Our hypothesis is supported. Based on the teachers' assessment, the accuracy of the explanation is increased by 15% when it is generated by assertion enhanced few-shot prompt than the traditional few-shot prompt. Assertion enhanced few-shot prompt could generate 67% accurate explanations, whereas traditional few-shot prompt could generate 52% accurate explanations.

**Table 1: Example of generated explanations after adding more demonstrations (Traditional Few-Shot Learning with more demonstrations ($k$ = 12)), adding assertions in the output (Traditional Few-Shot Learning with Assertions embedded into demonstrations) for test inputs**

| Test Input 1 |
|---|
| <$4v - 17 = -4v$, divide -4, Incorrect> |
| <two step with two variable terms, divide -ve, incorrect> → new added demonstration |

**Assertion Enhanced Few-Shot Learning**

because we can only divide if we have only one variable term in the entire equation isolated on one side on its own. Here we have two variable terms, 4v and -4v, on the left side of the equation. Therefore, we cannot divide by -4 at this step. We need to combine like terms first by adding 4v to both sides of the equation. This will result in $4v - 17 + 4v = -4v + 4v$.

**Traditional Few-Shot Learning ($k$ = 8)**

Because our goal is to isolate the variable v and -4 is multiplied with v. Therefore, we should divide by -4, not just 4. Here -4 is also called the coefficient. The coefficient is the number that is multiplied with the variable. Dividing by -4 will result in (4v-17)/-4 = (-4v)/-4.

**Traditional Few-Shot Learning with more demonstrations ($k$ = 12)** Because our goal is to isolate the variable v and -4 is multiplied with v. Therefore, we should divide by -4, not just 4. Here -4 is also called the coefficient. The coefficient is the number that is multiplied with the variable. Dividing by -4 will give us the correct solution.

**Traditional Few-Shot Learning with Assertions embedded into demonstrations**

Because our goal is to isolate the variable v and -4 is multiplied with v. The opposite of multiplication is division. However, when we divide by -4, we need to remember that dividing by a negative number changes the direction of the inequality sign. So, we should divide by -4 on both sides, resulting in (4v-17)/-4 = (-4v)/-4. Remember, we must divide if we have only one variable term in the entire equation on one side of "=" and one constant term on the other side of "="

| Test Input 2 |
|---|
| $<2x + 12 = -x$, divide 2, Incorrect> |
| <two step with two variable terms, divide +ve, incorrect> → already exists in demonstrations |

**Assertion Enhanced Few-Shot Learning**

because we have two variable terms in this equation, 2x and -x. We cannot divide when we have more than one variable term present in the equation. To isolate x, we need to get rid of one of the variable terms. We can do this by adding x to both sides of the equation. This will result in $2x + 12 + x = -x + x$, which simplifies to $3x + 12 = 0$

**Traditional Few-Shot Learning ($k$ = 8)**

Because our goal is to isolate the variable x and 2 is added with x. Therefore, we should subtract 2x first as it is the opposite of +2x. This would result in $2x + 12 - 2x = -x - 2x$.

**Traditional Few-Shot Learning with more demonstrations ($k$ = 12)** Because our goal is to isolate the variable x and 2 is multiplied with x. Therefore, we should divide by 2, not just 2. Here 2 is also called the coefficient. Coefficient is the number that is multiplied with the variable. So, the correct solution step would be to divide both sides of the equation by 2.

**Traditional Few-Shot Learning with Assertions embedded into demonstrations**

Because our goal is to isolate the variable x and 2 is added with x. The opposite of +2 is -2. So we must subtract 2 to isolate x on its own. This will result in $2x + 12 - 2\ = -x - 2$. Remember, subtracting a positive has the same result as adding the negation of that term. Therefore, the correct solution step would be to subtract 2, not divide by 2.

| Test Input 3 |
|---|
| $<10 = -6 - 4y$, add 6, Correct> |
| <two step with one variable term, add +ve, correct> → new added demonstration |

**Assertion Enhanced Few-Shot Learning**

Because our goal is to isolate the variable y and -6 is subtracted from -4y. The opposite of -6 is +6. So we must add 6 to isolate y on its own. This will result in $10 + 6\ = -4y + 6$.

**Traditional Few-Shot Learning ($k$ = 8)**

Because our goal is to isolate the variable y and -4y is subtracted from -6. The opposite of -4y is +4y. So we must add 4y to isolate y on its own. This will result in $10 + 6\ = -6 + 4y$.

**Traditional Few-Shot Learning with more demonstrations ($k$ = 12)** Because our goal is to isolate the variable y and -4y is subtracted from -6. The opposite of -4y is +4y. So we must add 4y to isolate y on its own. This will result in $10 + 6 = -6 - 4y + 4y$.

**Traditional Few-Shot Learning with Assertions embedded into demonstrations**

Because our goal is to isolate the variable y and -4y is subtracted from -6. The opposite of -4y is +4y. So we must add 4y to isolate y on its own. This will result in $10 + 6\ = -6 + 4y$. Remember, adding a positive term has the same result as subtracting the negation of that term.

| Test Input 4 |
|---|
| $<6 + n = -2 - 4n$, add 4, Incorrect> |
| <three step, add +ve, incorrect> → new added demonstration |

**Assertion Enhanced Few-Shot Learning**

Because our goal is to isolate the variable n, we need to get rid of the -4n term on the right side of the equation. To do this, we should add 4n to both sides, not just 4. Adding 4n to both sides will result in $6 + n + 4n = -2 - 4n + 4n$.

**Traditional Few-Shot Learning ($k$ = 8)**

Because our goal is to isolate the variable n and 4 is added to n. The opposite of +4 is -4. So, we must subtract 4 to isolate n on its own. This will result in $6 + n - 4\ = -2 - 4n - 4$

**Traditional Few-Shot Learning with more demonstrations ($k$ = 12)** Because our goal is to isolate the variable n and 4 is added to n. Here we have -2 subtracted with 4n, we want to do the opposite of -4n which is +4n. So, we must subtract 4n to isolate n on its own.

**Traditional Few-Shot Learning with Assertions embedded into demonstrations**

Because our goal is to isolate the variable n and 4 is added to n. The opposite of +4 is -4. So we must subtract 4 to isolate n on its own. This will result in $6 + n - 4 = -2 - 4n - 4$. Remember, subtracting a positive term has the same result as adding the negation of that term.

| Test Input 5 |
|:---:|
| <$6 + n = -2 - 4n$, subtract -2, Correct> |
| <three step, subtract -ve, correct> → new added demonstration |

**Assertion Enhanced Few-Shot Learning**

Because our goal is to isolate the variable n, we need to get rid of the -2 on the right side of the equation. Subtracting -2 is the same as adding 2, so we can rewrite the equation as $6 + n + 2 = -4n$. This simplifies to $8 + n = -4n$.

**Traditional Few-Shot Learning ($k$ = 8)**

Because our goal is to isolate the variable n and -2 is subtracted from n. The opposite of -2 is +2. So we must add 2 to isolate n on its own. This will result in 6+n+2 = -2-4n+2.

**Traditional Few-Shot Learning with more demonstrations ($k$ = 12)** Because our goal is to isolate the variable n and -2 is added with -4n. To isolate -4n from -2, we must perform the opposite of +(-2) which is +2. So, adding 2 will get rid of one constant term on the right side leaving us with only one constant term on the left side.

**Traditional Few-Shot Learning with Assertions embedded into demonstrations**

Because our goal is to isolate the variable n and -2 is subtracted from n. The opposite of -2 is +2. So we must add 2 to isolate n on its own. This will result in $6 + n + 2 = -2 - 4n + 2$. Remember, adding a positive term has the same result as subtracting the negation of that term.

The most commonly studied educational domain used to assess few-shot capabilities of LLM is arithmetic reasoning (like solving math word problems). The state-of-the-art accuracy for this task was 57% [15]. We hypothesize that as the complexity of the domain increased, the task accuracy would decrease, particularly when relying solely on a few-shot demonstration approach. This hypothesis aligns with the reported accuracy trends in [15] for math word problem solving, commonsense question answering and letter concatenation reasoning task were 57%, 80% and 100% respectively. Therefore, we argue that 15% increase in accurate explanations, achieved solely through prompting without the need for fine-tuning, is highly significant in the educational domain.

## 8.2 RQ2: Does Assertion Enhanced Few-Shot Learning enhance the quality of knowledge-building explanations generated by LLM, as assessed by teachers?

Our hypothesis is supported. Based on the teachers' assessment, the quality score of assertion enhanced few-shot prompt was higher than traditional few-shot prompt. The individual analysis of survey statements, which were aggregated to calculate quality score, revealed that the difference in the quality score can be attributed to the capability of assertions to generate explanations that focus on the optimal step, remain relevant to the input, and incorporate appropriate concept terms. However, both prompts did not exhibit significant differences in terms of reflecting a genuine comprehension of the concept based on how it is utilized within the explanation and no difference in terms of logical correctness. Our hypothesis is that the manual incorporation of assertions may not cover all the assertions required to attain a comprehensive understanding of concept terms. This implies that automated inclusion of assertions, especially in intricate domains, could be essential—an avenue we plan to investigate in future research.

### 8.3 Design Guideline 1: Assertion Enhanced Few-Shot Learning when accompanied with appropriate demonstrations generates better quality explanations

The distribution of teacher ratings for explanations generated from both the prompts on test inputs that closely resemble to those provided in the demonstrations (Figure 3) reveals that relying solely on the chain-of-thought patterns in the demonstrations was insufficient to generate highly quality explanations. Inclusion of assertions helped LLM to replicate the appropriate concept terms in the explanation of these similar test inputs, further improving the teacher ratings across all metrics. This finding indicates that assertions when accompanied with appropriate demonstrations help the generation of high-quality knowledge-building explanations.

### 8.4 Design Guideline 2: Assertions and the chain-of-thought text in the output must be consistent in terminology for the same concepts

In cases where the LLM had not encountered the test input within their demonstrations, we observed two contrasting outcomes in the generated explanations. For some unseen test inputs where LLM struggled to generate high-quality explanations solely conditioning on the demonstrations, the inclusion of assertions bridged the gap and resulted in improved explanations as evident from our interaction plot. Conversely, when the traditional few-shot prompt could independently generate higher-quality explanations, the addition of assertions did not yield a significant improvement. This suggests that the presence of some asserted facts might conflict with the generalized patterns learned from the demonstrations, potentially leading to suboptimal performance. This underscores the importance of incorporating assertions in a manner that aligns with the terminology used in the chain-of-thought to help LLM replicate accurate concepts.

## 9 LIMITATIONS AND FUTURE WORK

Our study has several limitations that should be considered. Firstly, in the design of our demonstrations, we did not account for the potential impact of changing the order of the demonstrations, which has been shown to substantially influence text generation in LLMs [33].

Secondly, the selection of the inputs in the demonstrations was somewhat arbitrary, and the results may not be entirely reproducible with a different set of few-shot demonstrations. Additionally, we did not specifically investigate the optimal number of demonstrations required for our domain, and this remains an area for further exploration. Similarly, in our ablation study, we directly increased the number of demonstrations to an arbitrary level. It is unclear whether the inclusion of assertions is equivalent to having a different optimal number of demonstrations, and this warrants further investigation.

Lastly, our approach for the incremental addition of assertions may not be feasible in domains such as medicine or other fields with a high volume of assertive facts. In such cases, it may be necessary to divide the task into simpler sub domains or explore the potential for automated assertion inclusion, which is part of our future research agenda.

## 10 CONCLUSION

In this work, we proposed a new prompting technique, which we termed *Assertion Enhanced Few-Shot Learning*. This technique combines the strengths of few-shot demonstrations, chain-of-thought reasoning, and the integration of domain-specific knowledge through assertions. We also explored and evaluated our proposed technique with traditional few-shot only technique and found significant improvements in both the accuracy and overall quality of the generated knowledge-building explanations, as assessed by teachers.

One noteworthy finding from our study was the impact of the placement of assertions within the few-shot demonstrations. We discovered that simply appending assertions to the chain-of-thought reasoning in the demonstrations led to a deterioration in the quality of the explanations. This observation reinforced our deliberate design decision to position the assertions at the end of the few-shot demonstrations. This arrangement not only enhanced the performance of the model but also proved to be a practical choice for instructional designers, educators, and domain experts. Placing assertions separately at the end of the demonstrations allows for easier integration and ensures that these valuable pieces of domain-specific knowledge are consistently applied across various inputs.

Our research findings hold potential for educational technology developers seeking to improve their systems, enabling them to provide high-quality, contextually relevant explanations. As we continue to explore the potential automation of assertion inclusion and its application in complex domains, our work contributes insights and tools for the education technology communities, ultimately enhancing the learning experiences of students and users worldwide.

## A   APPENDIX

You are an accomplished student skilled in solving <domain name>.

Your task is to provide a correct knowledge-building explanation using only 4 sentences when presented with the problem state, the solution step and information on whether the solution step is correct or incorrect.

A few examples of knowledge-building explanation are provided below delimited by triple quotes.
'''
<context>, <output>

<context>, <output>

.
.

<context>, <output>
'''

    <context>, <output>

**Table 2: A general format of Traditional Few-Shot Prompt**

<**Traditional Few-Shot Prompt**> +

A few facts about this domain that you must assert while generating a knowledge-building explanation:

    <domain assertions>

You will be heavily penalized if your generated knowledge-building response contradicts with the facts of the domain.

**Table 3: A general format of Assertion Enhanced Few-Shot Prompt**

You are an accomplished middle school student skilled in solving linear algebraic equations.
Your task is to provide a correct knowledge-building response using only 4 sentences when presented with the equation state, the solution step and information on whether the solution step is correct or incorrect.

A few examples of knowledge-building responses are provided below delimited by triple quotes.
'''
Given equation state: j-6=-9, given solution step: add 6, correctness of the given solution step: correct, knowledge-building response: because our goal is to isolate the variable j and 6 is subtracted from j. The opposite of -6 is +6. So we must add 6 to isolate j on its own. This will result in j-6+6 = -9+6.\

Given equation state: x+10=18, given solution step: subtract 18, correctness of the given solution step: incorrect, knowledge-building response: because our goal is to isolate the variable x and 18 is not related with x. Here we have 10 added with x, we want to do the opposite of +10 which is -10. So we must subtract 10 to isolate x on its own.\

Given equation state: -2y=8, given solution step: divide 2, correctness of the given solution step: incorrect, knowledge-building response: because our goal is to isolate the variable y and -2 is multiplied with y. Therefore, we should divide by -2 not just 2. Here -2 is also called the coefficient. Coefficient is the number that is multiplied with variable.\

Given equation state: 8v=16, given solution step: divide 8, correctness of the given solution step: correct, knowledge-building response: because our goal is to isolate the variable v and 8 is multiplied with v. The opposite of multiplication is division. So, we divide by 8 on both sides.\

Given equation state: 9-2z=6, given solution step: subtract 9, correctness of the given solution step: correct, knowledge-building response: because 9 is added with -2z. 9 is added here because there is no + or - sign infront of it. Therefore, we should do the opposite of +9 which is -9. This would result in 9-2z-9=6-9.\

Given equation state: 3z=9-z, given solution step: add z, correctness of the given solution step: correct, knowledge-building response: because we have two variable terms in this equation. add z is the opposite of -z. It will help us get rid of -z leaving us with only one variable term on one side of the equal sign.\

Given equation state: 2x=8x+10, given solution step: divide 2, correctness of the given solution step: incorrect, knowledge-building response: because we can only divide if we have only one variable term in the entire equation isolated on one side on its own. Therefore, we should subtract 8x first as it is the opposite of +8x.\

Given equation state: 11u+10=12u+17, given solution step: subtract 11u, correctness of the given solution step: correct, knowledge-building response: because subtract 11u will get rid of the variable term 11u on the left side leaving us with only one variable term on the right side.\
'''

Given equation state: <equation>, given solution step: <solution_step>, correctness of the given solution step: <correctness>, knowledge-building response: <generate your response here>

Table 4: The Traditional Few-Shot Prompt used in this work

<Traditional Few-Shot Prompt> +

A few facts about solving linear algebraic equation domain that you must assert while generating a knowledge-building response:
   - variable term must have a letter attached with a number.
   - The number attached with variable is called coefficient.
   - we can add or subtract a fraction directly if the fraction is related to the variable term.
   - We must divide if we have only one variable term in the entire equation on one side of "=" and one constant term on the other side of "=".
   - We must not divide when we have more than one variable term present in the equation.
   - When we have two variable terms and one constant term, we must get rid of the variable term from the side that has more terms.
   - Adding a positive term has the same result as subtracting the negation of that term.
   - Subtracting a positive has the same result as adding the negation of that term.

You will be heavily penalized if your generated knowledge-building response contradicts with the facts of the domain.

Table 5: The Assertion Enhanced Few-Shot Prompt used in this work

**[1]** [1] Given equation state: j-6=-9, given solution step: add 6, correctness of the given solution step: correct, knowledge-building response: because our goal is to isolate the variable j and 6 is the constant term subtracted from j. The opposite of -6 is +6. So we must add 6 to isolate j on its own. This will result in j-6+6 = -9+6.

**[2]** [2] Given equation state: x+10=18, given solution step: subtract 18, correctness of the given solution step: incorrect, knowledge-building response: because our goal is to isolate the variable x and 18 is not related with x. Here we have 10 added with x, we want to do the opposite of +10 which is -10. So we must subtract 10 to isolate x on its own.

**[3]** [3] Given equation state: -2y=8, given solution step: divide 2, correctness of the given solution step: incorrect, knowledge-building response: because our goal is to isolate the variable y and -2 is multiplied with y also known as coefficient. Therefore, we should divide by -2 not just 2. This will result in $\frac{-2y}{-2} = \frac{8}{-2}$.

→ [4] Given equation state: 15=3s-9, given solution step: add 9, correctness of the given solution step: correct, knowledge-building response: because our goal is to isolate 3s and 9 is subtracted from 3s. The opposite of -9 is +9. So we must add 9 to isolate 3s on its own. This will result in 15+9=3s-9+9.

**[4]** [5] Given equation state: 8v=16, given solution step: divide 8, correctness of the given solution step: correct, knowledge-building response: because our goal is to isolate the variable v and 8 is multiplied with v. The opposite of multiplication is division. So, we divide by 8 on both sides.

**[5]** [6] Given equation state: 9-2z=6, given solution step: subtract 9, correctness of the given solution step: correct, knowledge-building response: because 9 is added with -2z. 9 is added here because there is no + or - sign in front of it. Therefore, we should do the opposite of +9 which is -9. This would result in 9-2z-9=6-9.

**[6]** [7] Given equation state: 3z=9-z, given solution step: add z, correctness of the given solution step: correct, knowledge-building response: because we have two variable terms in this equation. add z is the opposite of -z. It will help us get rid of -z leaving us with only one variable term on one side of the equal sign.

**[7]** [8] Given equation state: 2x=8x+10, given solution step: divide 2, correctness of the given solution step: incorrect, knowledge-building response: because we can only divide if we have only one variable term in the entire equation isolated on one side on its own. Therefore, we should subtract 8x first as it is the opposite of +8x.

→ [9] Given equation state: r+2=-4r, given solution step: divide -4, correctness of the given solution step: incorrect, knowledge-building response: because we can only divide if we have only one variable term in the entire equation isolated on one side on its own. Therefore, we should subtract r first as it is the opposite of +r.

**[8]** [10] Given equation state: 11u+10=12u+17, given solution step: subtract 11u, correctness of the given solution step: correct, knowledge-building response: because subtract 11u will get rid of the variable term 11u on the left side leaving us with only one variable term on the right side.

→ [11] Given equation state: g-10=1-8g, given solution step: add 1, correctness of the given solution step: incorrect, knowledge-building response: because 1 is added with -8g. To isolate -8g from 1, we must perform the opposite of +1 which is -1. So, subtract 1 will get rid of one constant term on the right side leaving us with only one constant term on the left side.

→ [12] Given equation state: 2-3v=4+2v, given solution step: subtract -3v, correctness of the given solution step: correct, knowledge-building response: because subtract -3v will get rid of the variable term -3v on the left side leaving us with only one variable term on the right side.

Table 6: Existing demonstrations (shown with bold numbers) in the Traditional Few-Shot prompt with the newly added demonstrations (shown as → with underlined numbers) for the ablation study